\documentclass[11pt,letterpaper]{article}
\usepackage{emnlp2016}
\usepackage{times}
\usepackage{latexsym}
\usepackage{epstopdf}
\usepackage{graphicx}
\usepackage{color}
\usepackage{booktabs}
\usepackage{multirow}
\usepackage{url}

\usepackage{tikz}

\emnlpfinalcopy

\def\emnlppaperid{1103}

\title{Bag of Tricks for Efficient Text Classification} 

\author{
Armand Joulin~~~~~ 
Edouard Grave~~~~~
Piotr Bojanowski~~~~~
Tomas Mikolov \\
Facebook AI Research\\
\texttt{\{ajoulin,egrave,bojanowski,tmikolov\}@fb.com}
}

\date{}

\begin{document}

\maketitle

\begin{abstract}
This paper explores a simple and efficient baseline for text classification.
Our experiments show that our fast text classifier~\texttt{fastText} is often
on par with deep learning classifiers in terms of accuracy, and many orders of
magnitude faster for training and evaluation.  We can train~\texttt{fastText}
on more than one billion words in less than ten minutes using a standard
multicore~CPU, and classify half a million sentences among~312K classes in less
than a minute.  
\end{abstract}

\section{Introduction}

Text classification is an important task in Natural Language Processing with many
applications, such as web search, information retrieval, ranking and
document classification~\cite{deerwester1990indexing,pang2008opinion}.
Recently, models based on neural networks have become increasingly
popular~\cite{kim2014convolutional,zhang2015text,conneau2016}.
While these models achieve very good performance in practice, they tend to be
relatively slow both at train and test time, limiting their use on very large
datasets.

Meanwhile, linear classifiers are often considered as strong baselines for text
classification
problems~\cite{joachims1998text,mccallum1998comparison,fan2008liblinear}.
Despite their simplicity, they often obtain state-of-the-art performances if
the right features are used~\cite{wang2012baselines}.  They also have the
potential to scale to very large corpus~\cite{agarwal2014reliable}.

In this work, we explore ways to scale these baselines to very large corpus
with a large output space, in the context of text classification. Inspired by
the recent work in efficient word representation
learning~\cite{mikolov2013efficient,levy2014neural}, we show that
linear models with a rank constraint and a fast loss approximation can train on a billion
words within ten minutes, while achieving performance on par with the state-of-the-art.
We evaluate the quality of our approach \texttt{fastText}\footnote{\url{https://github.com/facebookresearch/fastText}} on two
different tasks, namely tag prediction and sentiment analysis.



\section{Model architecture}

A simple and efficient baseline for sentence classification is to represent
sentences as bag of words~(BoW) and train a linear classifier,~e.g., a logistic
regression or an~SVM~\cite{joachims1998text,fan2008liblinear}.
However, linear classifiers do not share parameters among features and classes.
This possibly limits their generalization in the context of large output space
where some classes have very few examples.
Common solutions to this problem are to factorize the linear classifier into low rank
matrices~\cite{schutze1992dimensions,mikolov2013efficient} or to use multilayer neural
networks~\cite{collobert2008unified,zhang2015character}. 

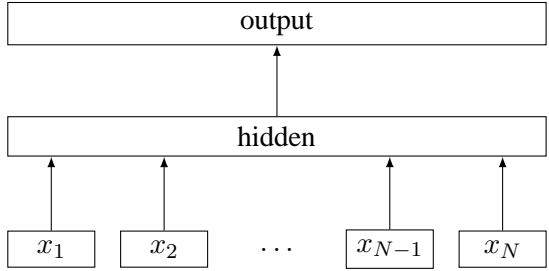
\begin{figure}
\centering
\begin{tikzpicture}[
  xblock/.style = {align=center, anchor=west, text width=.9cm},
  layer/.style  = {draw, align=center, anchor=west, text width=6.9cm},
  arrow/.style  = {draw, -latex}
]
\node[draw, xblock] (x1)   at (0,0) {$x_1$};
\node[draw, xblock] (x2)   at (1.5,0) {$x_2$};
\node[xblock]       (dots) at (3,0) {$\dots$};
\node[draw, xblock] (xn1)  at (4.5,0) {$x_{N-1}$};
\node[draw, xblock] (xn)   at (6,0) {$x_N$};
\node[layer]        (h)    at (0,1.5) {hidden};
\node[layer]        (o)    at (0,3) {output};
\draw[arrow] (x1.north)  -- (x1|-h.south);
\draw[arrow] (x2.north)  -- (x2|-h.south);
\draw[arrow] (xn1.north) -- (xn1|-h.south);
\draw[arrow] (xn.north)  -- (xn|-h.south);
\draw[arrow] (h.north)   -- (o.south);
\end{tikzpicture}
\caption{Model architecture of \texttt{fastText} for a sentence with $N$ ngram features $x_1,\dots,x_N$. The features
are embedded and averaged to form the hidden variable.}
\label{fig:model}
\end{figure}

Figure~\ref{fig:model} shows a simple linear model with rank constraint. The first
weight matrix~$A$ is a look-up table over the words. The
word representations are then averaged into a text representation, 
which is in turn fed to a linear classifier. The text representation
is an hidden variable which can be potentially be reused. This
architecture is similar to the cbow model of~\newcite{mikolov2013efficient},
where the middle word is replaced by a label.
We use the softmax function~$f$ to compute the probability distribution over
the predefined classes. For a set of~$N$ documents, this leads to minimizing
the negative log-likelihood over the classes: 
\begin{eqnarray*}
-\frac{1}{N} \sum_{n=1}^N y_n \log( f (BAx_n)),
\end{eqnarray*}
where~$x_n$ is the normalized bag of features of the~$n$-th document,~$y_n$ the label,~$A$ and~$B$ the weight matrices.
This model is trained asynchronously on multiple CPUs using stochastic gradient descent and a linearly decaying
learning rate.

\subsection{Hierarchical softmax}

When the number of classes is large, computing the linear classifier
is computationally expensive. More precisely, the computational complexity
is~$O(kh)$ where~$k$ is the number of classes and~$h$ the dimension of the text
representation.
In order to improve our running time, we use a hierarchical
softmax~\cite{goodman2001classes} based on the Huffman coding tree~\cite{mikolov2013efficient}.
During training, the computational complexity drops to~$O(h\log_2(k))$. 

The hierarchical softmax is also advantageous at test time when searching for
the most likely class. Each node is associated with a probability that is the
probability of the path from the root to that node.
If the node is at depth~$l+1$ with parents~$n_1,\dots,n_{l}$, its probability is
$$P(n_{l+1}) = \prod_{i=1}^l P(n_i). $$
This means that the probability of a node is always lower than the one of its parent.
Exploring the tree with a depth first search and tracking
the maximum probability among the leaves allows us to discard any branch
associated with a small probability. In practice, we observe a reduction of the
complexity to~$O(h\log_2(k))$ at test time. This approach is further
extended to compute the~$T$-top targets at the cost of~$O(\log(T))$, using a binary heap.

\subsection{N-gram features}

Bag of words is invariant to word order but taking explicitly this order into
account is often computationally very expensive. Instead, we use a bag of n-grams
as additional features to capture some partial information about
the local word order. This is very efficient in practice while
achieving comparable results to methods that explicitly use the
order~\cite{wang2012baselines}.

We maintain a fast and memory efficient mapping of the n-grams by using
the~\emph{hashing trick}~\cite{weinberger2009feature} with the same hashing
function as in~\newcite{mikolov2011strategies} and~10M bins if we only used
bigrams, and~100M otherwise. 

\begin{table*}[t]
\centering
\small
\begin{tabular}{@{\hspace{3pt}}l@{\hspace{3pt}}ccccccccc}
\toprule
Model && AG & Sogou & DBP & Yelp P. & Yelp F. & Yah. A. & Amz. F. & Amz. P. \\
\midrule
BoW~\cite{zhang2015character}          && 88.8 & 92.9 & 96.6 & 92.2 & 58.0 & 68.9 & 54.6 & 90.4 \\
ngrams~\cite{zhang2015character}       && 92.0 & 97.1 & 98.6 & 95.6 & 56.3 & 68.5 & 54.3 & 92.0 \\
ngrams TFIDF~\cite{zhang2015character} && 92.4 & 97.2 & 98.7 & 95.4 & 54.8 & 68.5 & 52.4 & 91.5 \\
char-CNN~\cite{zhang2015text}          && 87.2 & 95.1 & 98.3 & 94.7 & 62.0 & 71.2 & 59.5 & 94.5 \\
char-CRNN~\cite{xiao2016efficient}     && 91.4 & 95.2 & 98.6 & 94.5 & 61.8 & 71.7 & 59.2 & 94.1 \\
VDCNN~\cite{conneau2016}               && 91.3 & 96.8 & 98.7 & 95.7 & 64.7 & 73.4 & 63.0 & 95.7 \\
\midrule
\texttt{fastText}, $h=10$             && 91.5 & 93.9 & 98.1 & 93.8 & 60.4 & 72.0 & 55.8 & 91.2 \\
\texttt{fastText}, $h=10$, bigram     && 92.5 & 96.8 & 98.6 & 95.7 & 63.9 & 72.3 & 60.2 & 94.6 \\
\bottomrule
\end{tabular}
\caption{Test accuracy [\%] on sentiment datasets.
\texttt{FastText} has been run with the same parameters for all the datasets. 
It has $10$ hidden units and we evaluate it with and without bigrams.
For char-CNN, we show the best reported numbers without data augmentation. 
}\label{tab:sent_res}
\end{table*}

\begin{table*}[t]
\centering
\small
\begin{tabular}{@{\hspace{3pt}}cc@{\hspace{3pt}}cccccccc}
\toprule
&& \multicolumn{2}{c}{\newcite{zhang2015text}} && \multicolumn{3}{c}{\newcite{conneau2016}} && \texttt{fastText} \\ 
\cmidrule(l){3-4}\cmidrule(l){6-8}
&& small char-CNN & big char-CNN && depth=9 & depth=17 & depth=29 && $h=10$, bigram \\ 
\midrule
AG      && 1h & 3h && 24m  & 37m  & 51m  && 1s  \\
Sogou   && - & -   && 25m  & 41m  & 56m  && 7s \\
DBpedia && 2h & 5h && 27m  & 44m  & 1h   && 2s  \\
Yelp P. && - & -   && 28m  & 43m  & 1h09 && 3s \\
Yelp F. && - & -   && 29m  & 45m  & 1h12 && 4s \\
Yah. A. && 8h & 1d && 1h   & 1h33 & 2h   && 5s \\
Amz. F. && 2d & 5d && 2h45 & 4h20 & 7h   && 9s \\
Amz. P. && 2d & 5d && 2h45 & 4h25 & 7h   && 10s \\
\bottomrule
\end{tabular}
\caption{Training time for a single epoch on sentiment analysis datasets compared to char-CNN and VDCNN.
  }\label{tab:sent_speed}
\end{table*}

\section{Experiments}

We evaluate~\texttt{fastText} on two different tasks.  First, we compare it to
existing text classifers on the problem of sentiment analysis.  Then, we
evaluate its capacity to scale to large output space on a tag prediction
dataset.  Note that our model could be implemented with the Vowpal Wabbit
library,\footnote{Using the options \texttt{--nn}, \texttt{--ngrams} and \texttt{--log\_multi}}
but we observe in practice, that our tailored implementation is at
least~2-5$\times$ faster.

\subsection{Sentiment analysis}

\paragraph{Datasets and baselines.}
We employ the same~8 datasets and evaluation protocol of~\newcite{zhang2015character}.
We report the n-grams and~TFIDF baselines from~\newcite{zhang2015character}, as
well as the character level convolutional model~(char-CNN)
of~\newcite{zhang2015text}, the character based convolution recurrent
network~(char-CRNN) of~\cite{xiao2016efficient} and the very deep convolutional
network~(VDCNN) of~\newcite{conneau2016}.
We also compare to~\newcite{tang2015document} following their evaluation protocol.
We report their main baselines as well as their two approaches based on recurrent
networks~(Conv-GRNN and LSTM-GRNN). 

\paragraph{Results.}
We present the results in~Figure~\ref{tab:sent_res}.
We use~10 hidden units and run~\texttt{fastText} for~5 epochs with a learning rate selected
on a validation set from~$\{$0.05,~0.1,~0.25,~0.5$\}$.
On this task, adding bigram information improves the performance by~1-4$\%$. 
Overall our accuracy is slightly better than char-CNN and char-CRNN and, a bit worse than~VDCNN.
Note that we can increase the accuracy slightly by using more n-grams, for example
with trigrams, the performance on Sogou goes up to~97.1$\%$. 
Finally, Figure~\ref{tab:tang_res} shows that our method is competitive with the methods presented
in~\newcite{tang2015document}. We tune the hyper-parameters on the validation set and observe
that using n-grams up to~5 leads to the best performance.
Unlike~\newcite{tang2015document},~\texttt{fastText} does not use pre-trained word embeddings, which
can be explained the 1$\%$ difference in accuracy.

\begin{table}[h]
\centering
\small
\begin{tabular}{@{\hspace{2pt}}l@{\hspace{2pt}}cccc}
\toprule
Model & Yelp'13 & Yelp'14 & Yelp'15 & IMDB \\
\midrule
SVM+TF & 59.8 & 61.8 & 62.4 & 40.5 \\
CNN & 59.7 & 61.0 & 61.5 & 37.5\\
Conv-GRNN & 63.7 & 65.5 & 66.0 & 42.5 \\
LSTM-GRNN & 65.1 & 67.1 & 67.6 & 45.3 \\
\midrule
\texttt{fastText}          & 64.2 & 66.2 & 66.6 & 45.2 \\
\bottomrule
\end{tabular}
\caption{Comparision with \protect\newcite{tang2015document}. The hyper-parameters
are chosen on the validation set. We report the test accuracy.}\label{tab:tang_res}
\end{table}

\begin{table*}[t]
\centering
\small
\begin{tabular}{p{20em}clcp{15em}}
\toprule
Input &~~& Prediction &~~& Tags \\
\midrule
taiyoucon 2011 digitals: individuals digital photos from the anime convention
taiyoucon 2011 in mesa, arizona. if you know the model and/or the character,
please comment.
&&
{\#}cosplay
&&
{\#}24mm {\#}anime {\#}animeconvention {\#}arizona {\#}canon {\#}con {\#}convention {\#}cos \textbf{{\#}cosplay} {\#}costume {\#}mesa {\#}play {\#}taiyou {\#}taiyoucon
\\ 
\midrule[0.02em]
2012 twin cities pride 2012 twin cities pride parade
&&
{\#}minneapolis
&&
{\#}2012twincitiesprideparade \textbf{{\#}minneapolis} {\#}mn {\#}usa 
\\
\midrule[0.02em]
beagle enjoys the snowfall 
&&
{\#}snow
&&
{\#}2007 {\#}beagle {\#}hillsboro {\#}january {\#}maddison {\#}maddy {\#}oregon \textbf{{\#}snow} 
\\
\midrule[0.08em]
christmas && {\#}christmas && {\#}cameraphone {\#}mobile
\\
\midrule[0.02em]
euclid avenue && {\#}newyorkcity && {\#}cleveland {\#}euclidavenue 
\\
\bottomrule
\end{tabular}
\caption{Examples from the validation set of YFCC100M dataset obtained with \texttt{fastText}
with $200$ hidden units and bigrams. We show a few correct and incorrect tag predictions.
}\label{tab:tag_ex}
\end{table*}

\paragraph{Training time.}
Both char-CNN and VDCNN are trained on a~NVIDIA Tesla~K40~GPU, while our
models are trained on a~CPU using~20 threads.
Table~\ref{tab:sent_speed} shows that methods using convolutions are several orders of
magnitude slower than~\texttt{fastText}. While it is
possible to have a 10$\times$ speed up for char-CNN by using more recent~CUDA
implementations of convolutions,~\texttt{fastText} takes less than a minute to
train on these datasets. The~GRNNs method of~\newcite{tang2015document} takes
around~12 hours per epoch on~CPU with a single thread.
Our speed-up compared to neural network based methods increases with the size
of the dataset, going up to at least a~15,000$\times$ speed-up.

\subsection{Tag prediction}

\paragraph{Dataset and baselines.}
To test scalability of our approach, further evaluation is carried on the~YFCC100M dataset~\cite{ni15} which consists
of almost~100M images with captions, titles and tags.
We focus on predicting the tags according to the title and caption~(we do not use the images).
We remove the words and tags occurring less than~100 times and split the data
into a train, validation and test set.  The train set contains~91,188,648
examples~(1.5B tokens). The validation has~930,497 examples and the test set~543,424. 
The vocabulary size is~297,141 and there are~312,116 unique tags.
We will release a script that recreates this dataset so that our numbers could be reproduced.
We report precision at~1.

We consider a frequency-based baseline which predicts the most frequent tag. We
also compare with Tagspace~\cite{weston2014tagspace}, which is a tag prediction
model similar to ours, but based on the~Wsabie model of
~\newcite{weston2011wsabie}.  While the Tagspace model is described using
convolutions, we consider the linear version, which achieves comparable
performance but is much faster.

\begin{table}[t]
\centering
\small
\begin{tabular}{@{\hspace{2pt}}l@{\hspace{2pt}}ccc}
\toprule
\multirow{2}[3]{*}{Model} & \multirow{2}[3]{*}{prec@1} & \multicolumn{2}{c}{Running time}\\
\cmidrule(l){3-4}
& &  Train & Test \\
\midrule
Freq. baseline           & 2.2  & -    & - \\
Tagspace, $h=50$         & 30.1 & 3h8  & 6h \\
Tagspace, $h=200$        & 35.6 & 5h32 & 15h \\
\midrule
\texttt{fastText}, $h=50$          & 31.2 & 6m40 & 48s \\
\texttt{fastText}, $h=50$, bigram  & 36.7 & 7m47 & 50s \\ 
\texttt{fastText}, $h=200$         & 41.1 & 10m34  & 1m29 \\
\texttt{fastText}, $h=200$, bigram & 46.1 & 13m38  & 1m37 \\ 
\bottomrule
\end{tabular}
\caption{Prec@1 on the test set for tag prediction on
YFCC100M. We also report the training time and test time.
Test time is reported for a single thread, while training uses 20 threads for both models.
}\label{tab:tag_res}
\end{table}

\paragraph{Results and training time.}
Table~\ref{tab:tag_res} presents a comparison of~\texttt{fastText} and the baselines.
We run~\texttt{fastText} for~5 epochs and compare it to~Tagspace for two sizes of the
hidden layer,~i.e.,~50 and~200.
Both models achieve a similar performance with a small hidden layer, but adding
bigrams gives us a significant boost in accuracy. At test
time, Tagspace needs to compute the scores for all the classes which makes it
relatively slow, while our fast inference gives a significant speed-up when the
number of classes is large~(more than~300K here). Overall, we are more than an order of magnitude
faster to obtain model with a better quality. The speedup of the test phase is
even more significant~(a~600$\times$ speedup). Table~\ref{tab:tag_ex} shows
some qualitative examples.

\section{Discussion and conclusion}

In this work, we propose a simple baseline method for text
classification. Unlike unsupervisedly trained word vectors from word2vec, our
word features can be averaged together to form good sentence representations.
In several tasks, \texttt{fastText} obtains performance on par with recently
proposed methods inspired by deep learning, while being much faster.  Although
deep neural networks have in theory much higher representational power than
shallow models, it is not clear if simple text classification problems such as
sentiment analysis are the right ones to evaluate them.  We will publish our
code so that the research community can easily build on top of our work.

\paragraph{Acknowledgement.} We thank Gabriel Synnaeve, Herv\'e G\'egou,
Jason Weston and L\'eon Bottou for their help and comments. We also thank
Alexis Conneau, Duyu Tang and Zichao Zhang for providing us with
information about their methods.

\bibliography{emnlp2016}
\bibliographystyle{emnlp2016}

\end{document}